# EXTRACTING ARABIC RELATIONS FROM THE WEB

Shimaa M. Abd El-salam[1], Enas M.F. El Houby[1], A.K. Al sammak[2] and T.A. El-shishtawy[2,3]

[1]Department of Systems & Information, Engineering Division, National Research Centre, Cairo, Egypt
[2]Department of Electrical Engineering, Faculty of Engineering (Shoubra), Benha University, Egypt
[3]Department of Information Systems, Faculty of Computers and Informatics, Benha University, Egypt

## ABSTRACT

*There is a vast amount of unstructured Arabic information on the Web, this data is always organized in semi-structured text and cannot be used directly. This research proposes a semi-supervised technique that extracts binary relations between two Arabic named entities from the Web. Several works have been performed for relation extraction from Latin texts and as far as we know, there isn't any work for Arabic text using a semi-supervised technique. The goal of this research is to extract a large list or table from named entities and relations in a specific domain. A small set of a handful of instance relations are required as input from the user. The system exploits summaries from Google search engine as a source text. These instances are used to extract patterns. The output is a set of new entities and their relations. The results from four experiments show that precision and recall varies according to relation type. Precision ranges from 0.61 to 0.75 while recall ranges from 0.71 to 0.83. The best result is obtained for (player, club) relationship, 0.72 and 0.83 for precision and recall respectively.*

## 1. INTRODUCTION

There has been a growing interest in techniques for automatically extracting information from text. Specifically, Information Extraction (IE) [1, 2] is a process which takes texts as input and put them in a structured format, unambiguous data as output. The major research directions of IE are Named Entity Recognition (NER) such as (person, organization, location and so on) and Relations Extraction (RE) among them. Relation extraction has become in the last few years an interesting research domain. It is a very important for several applications, such as Web mining, Information retrieval, Questions answering, Social relationship network and Automatic summarization. RE is the task of recognizing the assertion of a particular relationship between two or more entities in text. Most relation extraction system focus on binary relation such as (is-a, part-of). The functional relation can be explicit or implicit. The explicit relation is indicated by a special word or sequence of words in the text, the implicit relation is a relation what can be mined from the text using the context [8].

In this paper, the semi-supervised system is introduced to extract relation from Arabic text; which based on bootstrapping. Semi-supervised learning has become an important topic in computational linguistics. For many languages processing tasks including relation extraction, there is an abundance of unlabeled data, where labeled data is lacking and too expensive to create in large quantities, therefore making bootstrapping techniques is desirable. The input is a small seed instance relations $(e_1, e_2)$ and retrieved summaries from Google search engine as a source text. Each seed instance pair is used with the data source to generate patterns. Then, those patterns are used to extract new entity instances of the same relation as the seed instances. An instance with a high credibility score will be used as a seed instance in the next iteration, and then will be used to construct a structured relation set. This iterative process will be terminated if extracted instances reached to the number that determined by the user. For simplicity, only binary relations are taken into account in this paper and the target language is Arabic. The main features of this work are as follows:

- The work is the first semi-supervised system that extracts binary relation instances from Arabic text using open source web contents.
- The Google search engine summaries are used as a corpus resource.
- The system is generic; it can be used in different domains based on the type of seed instances.
- The system detects Named Entities automatically through analysis of extracted patterns. Therefore, the system doesn¡t need specially prepared Named Entities dictionaries.
- The syntactic analysis is made for Arabic text to increase the efficiency of the results.

The remaining of the paper is organized as follow: Section 2 introduces the previous works that were used to extract relations from text. Section 3 describes the challenges of the Arabic language. Section 4 describes components of the proposed system in details. Section 5 presents experiments and results. Section 6 contains conclusions.

## 2. RELATED WORK

In this section, methods that extract semantic relations from texts will be reviewed. There are several works have been done for RE in English text, some of them were reviewed in [3], also there are some works for RE in Chinese [5, 6, 7], and other works in Portuguese were reviewed in [4] and as far as we know, very few works have been done for RE from Arabic text, it started from 2010 till now such as [8, 9, 10, 11, 12, 13]. These works were grouped according to the used technique into four categories [4, 8] rule-based, supervised, unsupervised and semi-supervised methods.

**Rule-based methods** utilize predefined linguistic (syntactic and semantic) rules written manually to extract relationships based on part of speech information. It is very interesting for a restricted domain and has a good quality of analysis. The major drawback of this approach is the disability to perform well in dealing with a wide range or new domain data. This is due to two reasons: rules should be rewritten for different tasks or when the application is enlarged to different domains and finding rules manually is very hard and time-consuming [15]. Ben Hamadou et al., [8] extracted functional relations between Arabic person names and organizations for Arabic-named entities using the NOOJ platform; they applied their experiment on a journalistic test corpus. Belkacem and Badr [9] presented some aspects to identify entities and relationships in Arabic text. They translated the Arabic sentences into a first order logic. Then, they used entities rules and relationships rules for extracting entities and relationships. Fatma Ali et al., [10] developed a prototype for Arabic text spatial relations (topological relations, directional relations, and distance relations). It used a rule library which consists of a set of grammatical rules and lexico-syntactic patterns. These rules are applied to raw Arabic text to extract the spatial relations. The pattern-specific regular expressions were applied over the pre-processed sentences to verify that the pattern indeed occurs in the sentence.

**Supervised methods** relation extraction depends on a classification task and use large corpus such as Ines Boujelben et al. [12] presented a tool called "RelANE" that detect all the semantic binary relations. For each word in the sentence, it uses its morphological, contextual and semantic features of entity types. Mohammed G.H. and Qasem [17] proposed a methodology that can extract ontological relationships. It can extract semantic features of Arabic text; propose syntactic patterns of relationships among concepts, and a formal model of extracting ontological relations. It has been designed to analyze Arabic text using lexical semantic patterns of the Arabic language according to a set of features.

**Unsupervised methods** use clustering techniques and similarities between features or context words such as Hasegawa [18] assumed that pairs occurring in similar contexts share the same type of relation, they can be clustered together, then named entity recognition is applied to identify the entities of interest. Zhang M. et al., [19] proposed a tree-similarity-based unsupervised learning method to extract relations between named entities from a large raw corpus. They modified tree kernels on relation extraction to estimate the similarity between parse trees more efficiently. Then, the hierarchical clustering algorithm was used to group entity pairs into different clusters. Finally, each cluster is labeled by an indicative word and unreliable clusters are pruned out.

**Semi-supervised methods** (weakly supervised) take a sample of patterns or some target relation instances for the purpose to acquire more basics until discovering all target relations such as Brin [20] proposed relation extraction system DIPRE, which starts with a small set of seed facts for one or more relations of interest. Then it looked for a pattern in sources as indicators of facts. Finally, it utilized these patterns to extract new facts. Agichtein and Gravano [21] proposed a system called Snowball, which adopted the similar strategy with DIPRE. However, Snowball didn't use exact match, but a similarity function to group similar patterns instead. Snowball's flexible matching system allows for slight variations in token or punctuation, it used NER to identify all organization and location entities. Ravichandran and Hovy [22] extracted simple surface patterns for extracting binary relations from The Web. It started with only a few examples of (question, answer) pairs. It focused on efficiency issues for scaling relation extraction to terabytes of data. It gave good results on specific relations such as birthdates, however, it had low precision on generic ones like (is-a) relation and (part-of) relation. Pennacchiotti and Panel [23] proposed a system called Espresso for extracting binary semantic relations. It started with a small set of seed instances for a particular relation, the system learned lexical patterns and then used the Web to filter and expand the instances. Rozenfeld and Feldman [24] implemented SRES (Self-Supervised Relation Extraction System). SRES took input names of the target relations and the types of their arguments. It then used a large set of unlabeled documents downloaded from the Web in order to learn the extraction patterns. SRES was related to the KnowItAll system which was developed by Oren Etzioni et al. [25]. Chao Chen et al. [5] developed REV (Relation Extraction with Verification). It extracted relations in Chinese text from World Wide Web, it just required a seed instances with the form of (e1, e2, keyword) as input and World Wide Web was used as data source. Yang et al. [6] proposed a tuple refinement method based on relationship

keyword extension. It constructed a keyword list based on the diversity of relation to extend the entity relation keyword, and then crawl the co-occurrence of entity and relation keywords by the redundant web information; it saved the highest statistical value of candidate entities as the second entity based on the principle of proximity and the predefined entity type. David S. et al. [16] proposed a bootstrapping system for relation extraction based on word embedding. It achieved better results when used related words to find similar relationships than with similarities between weighted vectors. It extracted four types of relationships from a collection of newswire documents.

## 3. THE CHALLENGES OF ARABIC LANGUAGE

In fact, the extracting relation from Arabic text is not an easy task due to challenges related to the Arabic language. Arabic language [14] is a language which different from Indo-European languages syntactically, semantically and morphologically.

The research progress of Arabic relation extraction is relatively limited. This limitation might be due to the nature of the Arabic language beside the lack of available linguistic resources. Actually, the accessible corpora are not annotated with named entities, and the relations do not include a sufficient number of annotated examples can be exploited for learning approaches. Arabic is a Semitic language that presents interesting morphological and orthographic challenges that could complicate the extraction of relations between NEs. In addition to the problems that are related to Arabic NE recognition and relation extraction task poses some specific challenges listed in [8, 12] some of these challenges are:

1) Multiple relations may be occurred between the same NEs.
2) Implicit Relations: that is not directly recognized by words in the text. They are mined from the text using contextual elements, for example relation is Belong-to.
3) It is necessary to use the previous context of the relation to know the missing element involved in the relation.
4) Interference between implicit relation and discontinuity.
5) The omission of one element of the relation between named entities NE (NE1, R, ?) or (?, R, NE2) or (NE1, ?, NE2).
These challenges which are listed above should be considered to achieve an efficient system for extracting relations.

## 4. THE PROPOSED SYSTEM

In this section, the proposed semi-supervised system for Arabic relations extraction will be illustrated. The purpose of the proposed system is to extract instances of binary semantic relations that lie in the same line of the Arabic text. The system extracts relation between two named entities $(e_1, e_2)$; $e_1$ should be a person named entity and $e_2$ any other named entity. The input to the proposed system is a small seed instance of the target relation and Google search result summaries as the source text. The output is a list of instances relations, i/e., named entities and the relation between them. The algorithm of the proposed system is shown in Figure 1.

The proposed system is an iterative process; each iteration consists of two phases 1) Pattern Extraction, 2) Instances Extraction. Each phase will be introduced in details in the following subsections.

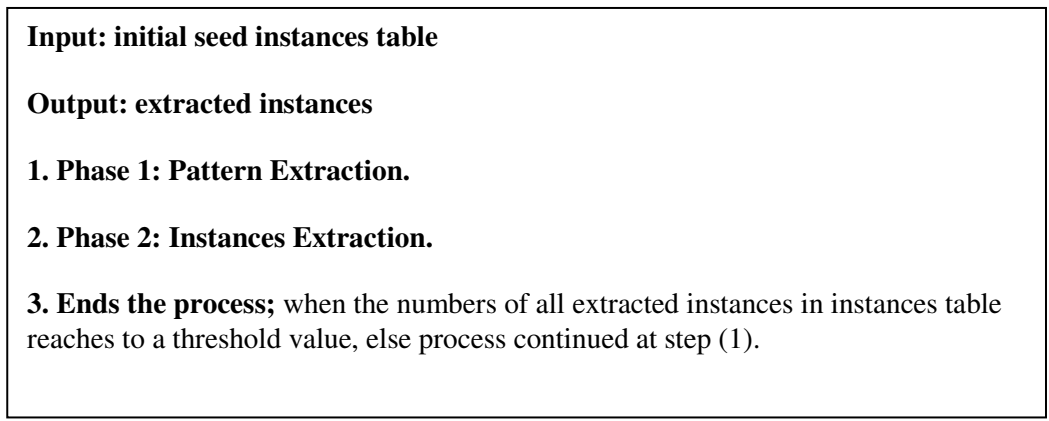

**Input: initial seed instances table**

**Output: extracted instances**

**1. Phase 1: Pattern Extraction.**

**2. Phase 2: Instances Extraction.**

**3. Ends the process;** when the numbers of all extracted instances in instances table reaches to a threshold value, else process continued at step (1).

Figure 1: The proposed algorithm that extracts relations from Arabic text.

## 4.1. Phase 1: Pattern Extraction

The input to this phase is examples of seed instance pairs, and the output is a set of their patterns. The algorithm of this phase is shown in figure 2.

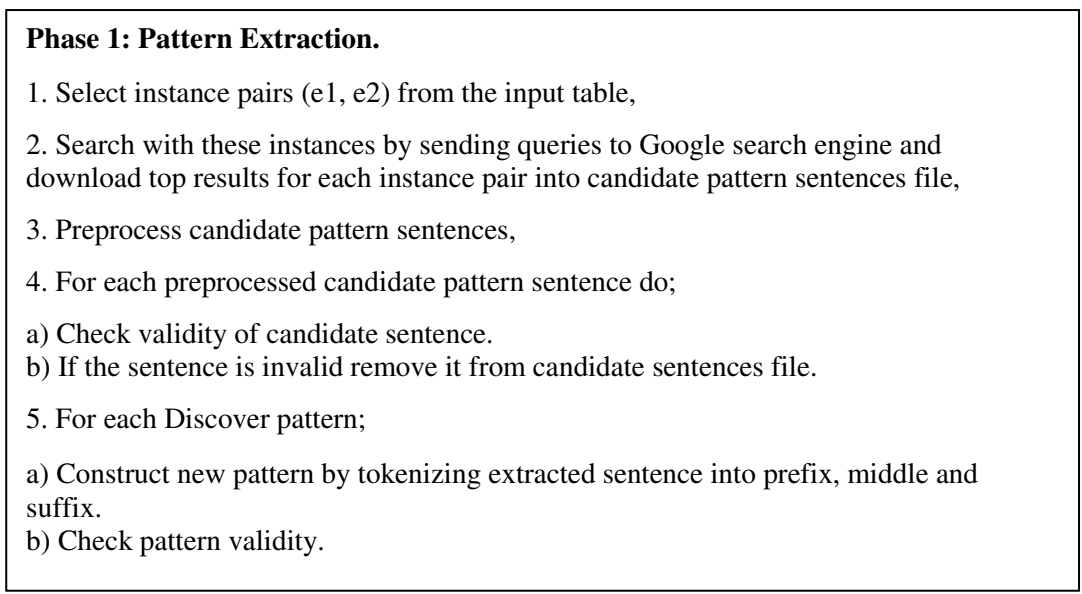

**Phase 1: Pattern Extraction.**

1. Select instance pairs (e1, e2) from the input table,

2. Search with these instances by sending queries to Google search engine and download top results for each instance pair into candidate pattern sentences file,

3. Preprocess candidate pattern sentences,

4. For each preprocessed candidate pattern sentence do;

a) Check validity of candidate sentence.
b) If the sentence is invalid remove it from candidate sentences file.

5. For each Discover pattern;

a) Construct new pattern by tokenizing extracted sentence into prefix, middle and suffix.
b) Check pattern validity.

Figure 2: Algorithm of Pattern Extraction

### 4.1.1. Input Instances

Similar to several semi-supervised methods such as DIPRE [20] and Snowball [21], the system starts with a small samples of the input initial seed instances ($e_1$, $e_2$) where $e_1$ and $e_2$ represent "named entity". In our proposed system, named entity can be person name, book title, organization name, etc. These initial instances pairs which stored in input table are used to start patterns extraction in the first iteration. The initial instances are given by the user and it can vary in size and domain to obtain more general patterns. From our analysis, it was observed that in the case of using a list of four initial seed is better than one or two or three seed to discover the many different patterns. Figure 3 shows that the greater the number of seed instances greater the number of extracted pattern, since the candidate patterns refer to all patterns with repetition and pattern extraction refer to the patterns without repetition. For example, using one seed instances it will give 13 candidate pattern 5 of them non-repeated patterns and then lifted slightly with 2, 3 seeds. But when using four seeds, extracted patterns will a significant increase increasingly so it was decided to use four seed instances as input.

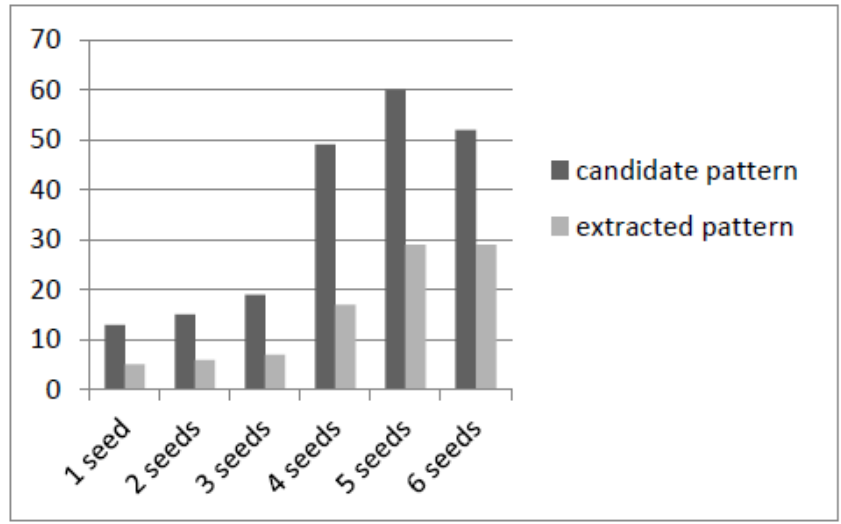

Figure 3: The relation between numbers of seeds and number of extracted patterns

## 4.1.2. Candidate Pattern Sentences

Candidate pattern sentences are Arabic sentences which are retrieved from the web when searching with instances. The proposed system searches for Arabic texts on the web with seed instances and using Google search engine. For each input instance $(e_1, e_2)$; it retrieves the first 20 top summaries results that contain the two terms $e_1$ and $e_2$. It was found that results above the 20 summaries contain unrelated or repeated results. Therefore, increasing the number collected above 20 summaries, does not affect the number of extracted patterns. So, choosing 20 top summaries for each pair is suitable. Then all these sentences are downloaded into a text file.

## 4.1.3. Preprocessing

The system deals with candidate pattern sentences line by line. Candidate pattern sentences string may contain non-Arabic letters and numbers which should be removed. The preprocessing task includes two steps: normalization and sentence segmentation. *Normalization* is the process in which the text is converted to UTF-8 character and any punctuation marks and non-Arabic letters are removed. Also, some Arabic letters are normalized such as" إ ", "أ ", and "آ"which are replaced with" ا ", and" ى"is replaced with "ي" and "ه" is replaced with " ة " and so on. Closed set of Arabic words related to internet web pages ex. ("مجانا" free, "موقع" site, "تحميل" download, and so on.) are removed. The second step is *Sentence Segmentation* which is a process of dividing a string of written language into sentence's components.

### 4.1.4. Sentence Extraction Validation

Candidate sentences may contain noisy results; therefore, checking a validity of the extracted sentences is very important to remove unrelated sentences. The sentence is marked as valid, if it suits the following criteria:-

i. Sentences must contain both entities ($e_1$, $e_2$) in the same line.
ii. A maximum number of words between the two entities should not exceed 3 words.

### 4.1.5. Pattern Discovery

#### *4.1.5.1. Pattern Construction*

The extracted valid sentences are analyzed and tokenized to prefix phrase, middle phrase, suffix phrase and order. They are analyzed according to an occurrence which consists of 6-dimension tuples:

**(*prefix, e₁, middle, e₂, suffix, order*)**

Where "prefix" consists of the n words preceding e1 if e1 was the first or preceding $e_2$ if $e_2$ was the first, "middle" is all words between the e1 and $e_2$, "$e_1$" is the first entity such as person name, "$e_2$" is the second entity such as organization name and "suffix" consists of the m words following $e_2$ if $e_2$ was the second entity or following $e_1$ if $e_1$ was the second. "Order" corresponds to the order of $e_1$ and $e_2$ in the text, where order is a Boolean value. The following criteria are considered to extract candidate patterns:

1. Verify that the maximum number of word in the middle between two entities is three words.
2. Check that number of words in each of prefix and suffix, if it is more than two words, then take only two words which before first entity $e_1$ as prefix and three words after second entity as a suffix.
3. If any of prefix, suffix and middle context haven't any word; set it = null.
4. Order takes "true" if $e_1$ occurred before $e_2$, otherwise, order takes "false".

#### *4.1.5.2. Pattern Validation*

The purpose of this stage is to remove duplicated extracted patterns. It depends on the similarity between the set of candidate patterns from the previous stage.

For instance, consider two occurrences $O_1$ and $O_2$ as follow:
$O_1$ (occurrence) = $p_1\{w_{11}, w_{12}\}, e_1, m_1\{w_{11}, w_{12}, w_{13}\}, e_2, s_1\{w_{11}, w_{12}\}$
$O_2$ (occurrence) = $p_2\{w_{21}, w_{22}\}, e_1, m_2\{w_{21}, w_{22}, w_{23}\}, e_2, s_2\{w_{21}, w_{22}\}$

Whereas "O" refers to the occurrence, "w" refers to word, "p" refers to a prefix, "m" refers to middle, and "s" refers to a suffix. The criteria of matching processes are as follows:

1. Verify that the *order* of occurrences is the same. If not, it is not possible to match these patterns, then save this pattern as extraction pattern with a number of repetition =1.

2. If prefix, middle words and order are the same but suffix contains different words in two occurrences; then save this pattern as extraction pattern with suffix = null, and the number of repetition equal number of matched patterns.

3. Verify prefix and middle words similarity. if not but there are some words similar, then calculate intersection between words in the *prefix* $p_i$ or *middle* $m_i$ for example if $w_{12}(p_1) = w_{22}(p_2)$ and if $w_{11}(m_1) = w_{23}(m_2)$; then a pattern will be P = $p(w_{12})$, $m(w_{11})$, $s(null)$ and order = 1 (because $e_1$ take places before $e_2$) and number of repetition =2.

## 4.2. Phase 2: Instance Extraction

In this phase, the system takes the patterns extracted from phase1 as input and retrieves a set of newly extracted instances that match these patterns as output. Algorithm of phase 2 is shown in figure 4.

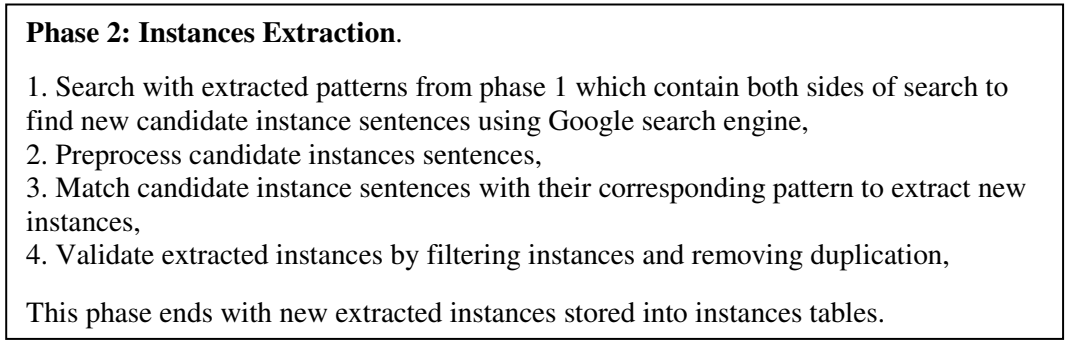

**Phase 2: Instances Extraction**.

1. Search with extracted patterns from phase 1 which contain both sides of search to find new candidate instance sentences using Google search engine,
2. Preprocess candidate instances sentences,
3. Match candidate instance sentences with their corresponding pattern to extract new instances,
4. Validate extracted instances by filtering instances and removing duplication,

This phase ends with new extracted instances stored into instances tables.

Figure 4: Algorithm of Instances Extraction

### 4.2.1. Candidate Instance Sentences

Candidate instance sentences are Arabic sentences which retrieved from the web when searching with pattern. The system ignores patterns that do not contain two parts of search. A query is generated for extracted patterns that contain both prefix and middle words. Prefix and middle words are used to retrieve the top 20 results for each pattern query. For each pattern, the query of search string takes the form:

Search pattern query: "+prefix words+$e_1$+middle words+ $e_2$+"

For example;                    "+مجموعة+*+المملوكة+لرجل+الاعمال+*+"

The retrieved candidate instance sentences for each searched pattern are collected together in a text file. Candidate instance sentences may contain non-Arabic letter and numbers. To overcome extracted noisy occurrences, the output of the search engine is validated against used patterns. It must be preprocessed similar to procedures illustrated in section 4.1.3 to remove any noise.

### 4.2.2. Instance Extraction

The system reads extracted sentences line by line from a text file. It use match method between each pattern with all extracted sentences to extract new instances pair ($e_1$, $e_2$). Each pattern consist of 5 dimension tuple < prefix, $e_1$, middle, $e_2$, suffix >.Then, the system cut the sentences that matched the pattern into 5 parts according to extracted pattern; where prefix, middle, and suffix are known, $e_1$ and $e_2$ are required. Finally, the system can detect all the potential instances. For example; extracted pattern is:

$$P= <(*?)> (رئيس جمهورية (*?) رئيس الدكتور (*?)>$$

This pattern matches sentences such as:

$$S_1= "....استقبل فخامة الرئيس الدكتوراكليل ظنين رئيس جمهورية جزر القمر المتحدة في مقَر اقامته بمكه..."$$

$$S_2= "....تسلم فخامة الرئيس الدكتوراولافور راغنار غريمسون رئيس جمهورية ايسلندا اوراق اعتماد سعادة السيد...."$$

Then the extracted instances are:

$$(ايسلندا اوراق اعتماد ,اولافور راغنار غريمسون) and (جزر القمر المتحدة ,اكليل ظنين)$$

### 4.2.3. Instance Validation

The goal of the instance validation stage is to filter the list of new extracted instances, keeping the correct instances and removing mistakes that may occur from patterns. The observations after extracting instances are that some instances are repeated many times. Moreover, in some instances name of a person may be duplicated with different format for the same second entity, because name of a person may be written by two or three words, so the goal of this step is to filter the list of extracted instances. Some constraints have been set to control new instance generation. Some filtering rules have been used instead of using Named Entities Recognition to minimize retrieving noise in extracted instances. The purpose of these rules is to evaluate and filter new instance to obtain correct instances which will be used in next iteration. These rules are:-

1. Remove one of the duplicated instances which are common in the two entities ($e_1$, $e_2$).
2. Remove one of the duplicated instances if two pairs of instances ($e_1$, $e_2$) and ($e_{1*}$, $e_{2*}$) have similar words such as $e_2$, $e_{2*}$ are the same words but words of $e_1$may exceed from$e_1$with other words, then they will be described by intersected part. For example: $e_1$"محمد نجيب"and $e_{1*}$"محمد نجيب عبدالله" will be represented as"محمد نجيب".
3. The maximum length of $e_1$is 3 words except if the third word is "عبد" or "بن"or "ابو". Only one more word will be considered in this case because it is incorrect to end a person name with "عبد" or "بن" or "ابو" according to the Arabic language. For example, after "عبد" one word such as ("الله", "الفتاح", "خليفه" and so on…) is allowed.
4. It is invalid instances if named entity formed from one word such as ("محمد", "من","في") except if the word starts with "ال"ex. "القرطبى","الطبرى" will be considered valid instances.

### 4.3. End Condition

The system will continue repeating the procedure and the results of extracted instances will be added to instances table at each iteration. After adding the results to the table of instances, the system will remove duplication for instances. The system uses only new instances as input to the next iteration in case of not reaching the end condition. Some other systems can be terminated when no new candidate pairs are extracted such as in [21], or when a human observer decides many pairs have been returned such as in [20].

In our proposed system, we decided to terminate the iterations when reached to the number that is satisfactory for the user or when total numbers of extracted instances in the table reaches to ≥ "a threshold value". This threshold value is set to 100 instances. We choose this value to terminate the iterations to be even easy for us to measure the performance of our system.

# 5. EXPERIMENTS AND RESULTS

## 5.1. Experimental Setting

In this section, we will talk about source of text that is used and our experiments. The proposed system uses World Wide Web and search engine as the source of text. In our experiments all of the sentences (candidate pattern sentences and candidate instance sentences) are extracted from Google search engine since the data source is open. Our system used Jsoup (Java Web Crawler) [16] for retrieving Google search results. Figure 5 shows the operation that fetches results from Google.

```java
for (int i = 0; count < 3; i = +10) {
try {
System.setProperty("http.proxyHost", "192.168.5.1");
System.setProperty("http.proxyPort", "1080");
doc = Jsoup.connect(googleURLString + searchString + "&start=" + i).
userAgent(userAgent).referrer("http://www.facebook.com") .timeout(5000).get();
Elements links = doc.select("li[class=g]");
for (Element link : links) {
Elements titles = link.select("h3[class=r]");
String title = titles.text();

Elements bodies = link.select("span[class=st]");
String body = bodies.text();

System.out.println("Title: " + title);
System.out.println("Body: " + body + "\n");
```

Figure 5: Jsoup operations that fetch results from Google

We have performed four relations to evaluate the performance of the proposed system. Four common relations which are the author-of (person, book) relation, president-of (person, country) relation, play-in (person, club) relation and CEO-of or chairman (person, company) relation are selected. Table 1 shows the sample of initial seed instance for each relation type. As described above, each system started with just 4 different seed instances for each relation to obtain the high coverage of extracted patterns.

Table 1: Sample of initial seed instances for each relation

| Relation | First entity  e₁ | Second entity e₂ |
|----------|------------------|------------------|
| (author, book) | مصطفى صادق الرافعي | وحي القلم |
| (president, country) | رجب طيب اردغان | تركيا |
| (player, club) | باسم مرسي | الزمالك |
| (chairman, company) | معتز الألفي | أمريكانا |

## 5.2. Assessment Methods

To evaluate the quality of our system we measured the performance by several methods. The first evaluation of extracted instances to measure their correctness, after that, the extracted pattern can be evaluated by its ability to extract correct instances.

### 5.2.1. Evaluating Instances

Instances can be evaluated by two methods the first is by human judges after that by calculating precision, recall and F-Measure.

We asked the human judge for each relation domain (political, sport, business and the literature). They should have good Arabic language skills and knowledge about these domains. We presented to them printed form that contains output extracted instances to judge and evaluate the correctness of new instances after terminating iterations. They evaluated manually using the internet. For each instance pair, we asked human judges to assign a score of 1 for correct, 0 for incorrect. After that, we computed numbers of instances (correct - incorrect) for each pattern to evaluate pattern in the next step. As different relation extract a different number of instances, in order to assess the experiment result, we use three common evaluation measures: Precision P, Recall R and F-Measure for all extracted instances of each relation. We measure the correctness of extracted instances according to existing correct ones using a recall metric, measuring the ability of our proposed methodology to extract instances with respect to all information using a precision metric, and finally, applied an F-measure that denotes the overall accuracy.

Given numbers of correctly classified extracted instances, denoted as Nc and actual number of correct instances concepts, denoted as A = Nc + Uc. Where Uc represents undetected instances, Uc were discovered manually by human judge for all text file and T represents the total number of extracted instances.

These metrics are defined as follows:

$$\text{Precision P} = \frac{Nc}{T}$$

$$\text{Recall R} = \frac{Nc}{A}$$

$$\text{F} - \text{Measure} = \frac{2 * P * R}{(P + R)}$$

### 5.2.2. Evaluating Patterns

Patterns can be evaluated by calculating confidence for each pattern with its extracted instances which were extracted from it. Patterns may be too general, thus they generate invalid instances. The probability of pattern in generating valid instances is calculated by the confidence of the extracted patterns for relations as Agichtein and Gravano [21]. The confidence of a pattern P is:-

$$\mathrm{Conf}\,(P) \;=\; \frac{positive\ instances}{positive\ instances + negative\ instances}$$

Where positive is the number of correct instances for P and negative is the number of incorrect instances. Table 2 shows that samples of confidence for some patterns.

For example, if this pattern P matches three Arabic sentences as shown:

p= "<$e_1$> قصص احدى هي <$e_2$>رواية";

$S_1$=" رواية رحلة ابن فطومه هي احدى قصص نجيب محفوظ الرائعة"

$S_2$=" رواية عودة الروح هي احدى قصص توفيق الحكيم"

$S_3$= " رواية بين القصرين هي احدى قصص الثلاثية"

The sentence $S_1$ generates instance pair (رحلة ابن فطومه, الرائعةنجيب محفوظ) and sentence $S_2$ generates (عودة الروح, توفيق الحكيم) which are correct instances, so the first 2 sentences are considered as positive instances. The third sentence $S_3$ generates instance (بين القصرين, الثلاثية) which are incorrect instances, so it is considered as a negative instance. $\mathrm{Conf}\,(P) = \frac{2}{2+1} = 0.66$. This result is satisfactory, where $\mathrm{Conf}\,(P) > 0.5$ and then extracted pattern is mostly reliable. To measure the reliability for patterns;

1. Calculate confidence for each pattern, if $\mathrm{Conf}\,(P) \geq 0.5$ it is a reliable pattern.
2. Calculate the number of instances that each pattern can extract or detect.
3. Check for new pattern if already exists in previous iterations. Then, this pattern is more reliable.

Table 2: Sample of patterns confidence for (author-of) relation

| pattern | Pos. Instances | Neg. Instances | Confidence |
|---|---|---|---|
| كتاب $< e_1 >$ ل $< e_2 >$ <br> Book$< e_2 >$ for $< e_1 >$ | 6 | 5 | 0.5 |
| رواية $< e_1 >$ ل $< e_2 >$ <br> Novel $< e_2 >$ for $< e_1 >$ | 7 | 1 | 0.8 |
| رواية$< e_1 >$ للكاتب $< e_2 >$ <br> novel $< e_2 >$ for writer $< e_1 >$ | 13 | 3 | 0.8 |
| كتاب $< e_1 >$ للشيخ $< e_2 >$ <br> Book $< e_2 >$ for Sheikh $< e_1 >$ | 8 | 2 | 0.8 |

## 5.3. Experimental Results and Analysis

In this section, we will represent the experimental results of the proposed system. As described above, each experiment started with just 4 different seed instances for each relation type. We will discuss each experiment in details.

**Experiment 1** is author-of (person, book) relation; in the first iteration, our proposed system extracted 23 patterns. Only 9 patterns of them extracted instances. These patterns detected 66 (author, title) pairs of instances, 2 instances were repeated from the input seeds. Due to total extracted instances did not reach the threshold value so it moved to next iteration and this output was added to the table of instances. New extracted instances will be used as input to next iteration. In the second iteration, the system extracted 92 patterns only 24 patterns of them detected instances. These patterns detected 145 pairs of instances. After adding these instances to the table of instances, We found that 25 instances are repeated. Then total extracted instances without repetition were 186 so system terminates iterations after the second iteration. After evaluating these instances, we found 52 incorrect instances and 45 instances were undetected by our system.

**Experiment 2** is president-of (person, country) relation; in the first iteration, our system extracted 10 patterns. Only 3 patterns of them extracted instances. These patterns detected 13 pairs of extracted instances 2 instances were repeated from the input seeds. After adding this output to the instances table, it was found that a total number of extracted instances did not reach a threshold value so it moved to next iteration. In the second iteration, the system extracted 30 patterns only 6 patterns of them extracted instances. These patterns detected 27 pairs of instances. After adding these instances to instances table, it was found that 3 instances were repeated. Then total extracted instances without repetition were 37 so our system went to next iteration due to total extracted instances did not reach a threshold value. In the third iteration, 38 patterns were extracted only 21 patterns of them detected instances. These patterns detected 74 instances, 63 pairs of instances were new instances. After adding these instances to the table of instances, it was found that 11 instances were repeated. Then total extracted instances without repetition were 100 so system terminates iterations after the third iteration. After evaluating these instances; we found that 39 incorrect instances and 14 instances were undetected by our system.

**Experiment 3** is play-in (person, club) relation; in the first iteration, our system extracted 17 patterns. Only 12 patterns of them extracted instances. These patterns detected 52 pairs of instances, in which 3 instances were repeated from the input seed. Due to total extracted instances did not get to a threshold value so it moved to next iteration and this output was added to the input table. In the second iteration, the system extracted 111 patterns only 56 patterns of them detected instances. These patterns detected 110 pairs of instances. After adding these instances to the table of instances, we found that 15 instances are repeated. Then total extracted instances without repetition were 147 so system terminates iterations after the second iteration. After evaluating these instances, we found that 41 incorrect instances and 21 instances were undetected by our system.

**Experiment 4** is CEO-of or chairman-of (person, company) relation; in the first iteration, our system extracted 14 patterns. Only 8 patterns of them extracted instances. These patterns detected 37 pairs of instances, in which 4 instances are repeated from the input seed. Due to total extracted instances did not reach a threshold value so it moved to next iteration and this output was added to the input table. In the second iteration, the system extracted 34 patterns only 18 patterns of them were detected instances. These patterns detected 96 pairs of instances. After adding these instances to the table of instances, we found that 18 instances were repeated. Then total extracted instances without repetition were 115 so system terminates iterations after the second iteration. After evaluating all these instances, we found that 29 incorrect instances and 25 instances were undetected by our system.

After applying our proposed system on four different relations types, we observed that the results of some patterns when searched in Google it gives English text results (summaries), we found that, the patterns which contain keywords about relation on prefix or middle give correct instances than the patterns that do not contain any keywords in prefix or middle. We found also that those patterns which are more general extracted incorrect instances and those patterns which are more specific extracted a small number of instances.

The results for all experiments are summarized in tables (3 and 4). It can be observed that three types of relations reached a threshold value of extracted instances after the second iteration and one relation "president-of" needed to the third iteration to reach this value as shown in table 3 Where "T" represents total number of extracted instances, "N" the number of newly extracted instances and R represents repeated instances in last iterations. R in the first iteration represents a number of instances repeated from input seed instances.

Table 3: Analysis of extracted instances in each iteration for all relations

| Extracted instances | 1st Iteration | | 2nd Iteration | | 3rd Iteration | | Total "T" |
|---|---|---|---|---|---|---|---|
| | R | N | R | N | R | N | |
| (Author, book) | 2 | 64 | 25 | 120 | - | - | 186 |
| (president, country) | 2 | 11 | 3 | 24 | 11 | 63 | 100 |
| (player, club) | 3 | 49 | 15 | 95 | - | - | 147 |
| (chairman, company) | 4 | 33 | 18 | 78 | - | - | 115 |

From table 4; **P$_{total}$** represents a total number of patterns extracted and **P$_{detect}$** represents a number of patterns that could extract instances. It was observed that P$_{detect}$ does not exceed one-third to half from the P$_{total}$.

Table 4: Analysis patterns for each relations

| Patterns | 1$^{st}$ Iteration | | 2$^{nd}$ Iteration | | 3$^{rd}$ Iteration | |
|---|---|---|---|---|---|---|
| | P$_{total}$ | P$_{detect}$ | P$_{total}$ | P$_{detect}$ | P$_{total}$ | P$_{detect}$ |
| (Author, book) | 23 | 9 | 92 | 24 | - | - |
| (president, country) | 10 | 3 | 30 | 6 | 38 | 21 |
| (player, club) | 17 | 12 | 111 | 56 | - | - |
| (chairman, company) | 14 | 8 | 34 | 18 | - | - |

By evaluating all extracted instances for each relation type after terminating the iterations we found that; the performance of proposed system nearly different from relation type to another as shown in table 5 where "NC" represents correct instances, "I" represents incorrect instances and "UC" represents undetected instances.

Table 5: Evaluation extracted instances for each relation

| Relation type | T | N$_C$ | I | U$_C$ |
|---|---|---|---|---|
| (Author, book) | 186 | 134 | 52 | 54 |
| (president, country) | 100 | 61 | 39 | 14 |
| (player, club) | 147 | 106 | 41 | 21 |
| (chairman, company) | 115 | 86 | 29 | 25 |

After evaluating each extracted patterns by measuring confidences, the average confidence for different iterations was calculated for four relations types as shown in Table 6. The average confidence ranges from 0.6 to 0.8. This means that extracted patterns are accurate.

Table 6: Average confidences for all relations types

| Patten Confidence | 1$^{st}$ Iteration | 2$^{nd}$ Iteration | 3$^{rd}$ Iteration | Avg. Confidence |
|---|---|---|---|---|
| (Author, book) | 0.8 | 0.6 | - | 0.7 |
| (president, country) | 0.3 | 0.8 | 0.7 | 0.6 |
| (player, club) | 0.7 | 0.8 | - | 0.8 |
| (chairman, company) | 0.8 | 0.7 | - | 0.8 |

Evaluating the performance of the proposed system for all relation types using precision, recall and f-measure is shown in Table 7. It can be observed that the best results have been achieved for (player, club) relation, whereas results for (president, country) relation are the lowest ones. The overall performance for the system is satisfactory.

Table 7: precision, Recall and F-measure for each relation type

| Relation type | Precision | Recall | F-Measure |
|---|---|---|---|
| (Author, book) | 0.72 | 0.71 | 0.71 |
| (president, country) | 0.61 | 0.81 | 0.69 |
| (player, club) | 0.72 | 0.83 | 0.77 |
| (chairman, company) | 0.75 | 0.77 | 0.76 |

## 6. CONCLUSIONS AND FUTURE WORKS

In this paper, we introduced a semi-supervised system to extract binary semantic relations for Arabic text from the Web. It only needed a few seed instances as input. Google search engine was used as a source of text. Our system depended on the pattern-based system. The system extracted patterns then searched with these patterns in Google search engine to extract new instances. These new instances were filtered to avoid a noise. Only new unrepeated instances are used as input instances for the next iteration. This iterative process was terminated after a total number of extracted instances reached a threshold value that determined by the user. We performed four experiments for four relations to evaluate the performance of the proposed system. Four common relations were the author-of (person, book) relation, president-of (person, country) relation, play-in (person, club) relation and CEO-of or chairman (person, company) relation. Our system was evaluated by four human judges to measure the correctness for extracted instances. The quality of patterns was evaluated after terminating the system by calculating confidence for each pattern with its extracted instances. After that, we measured the performance of the system by calculating Precision, Recall, and F-Measure for extracted instances for each relation type, the result was different from domain to domain. The system can be applied on other domains such as movies (director's name, film's name), music (singer, song), political (head's name, political party) and so on.

Our future work will focus on the improvement of rules of filtering to obtain corrected instances. And this work will be applied on another dataset to compare its results with results of Google search. Finally, we would like to extract complex relation.